# A Single Video Super-Resolution GAN for Multiple Downsampling Operators based on Pseudo-Inverse Image Formation Models

Santiago López-Tapia, Alice Lucas, Rafael Molina and Aggelos K. Katsaggelos.

*Abstract*—The popularity of high and ultra-high definition displays has led to the need for methods to improve the quality of videos already obtained at much lower resolutions. Current Video Super-Resolution methods are not robust to mismatch between training and testing degradation models since they are trained against a single degradation model (usually bicubic downsampling). This causes their performance to deteriorate in real-life applications. At the same time, the use of only the Mean Squared Error during learning causes the resulting images to be too smooth. In this work we propose a new Convolutional Neural Network for video super resolution which is robust to multiple degradation models. During training, which is performed on a large dataset of scenes with slow and fast motions, it uses the pseudo-inverse image formation model as part of the network architecture in conjunction with perceptual losses, in addition to a smoothness constraint that eliminates the artifacts originating from these perceptual losses. The experimental validation shows that our approach outperforms current state-of-the-art methods and is robust to multiple degradations.

*Index Terms*—Video, Super-resolution, Convolutional Neuronal Networks, Generative Adversarial Networks, Perceptual Loss Functions

## I. INTRODUCTION

The task of Super-Resolution (SR) consists of obtaining High-Resolution (HR) images from the corresponding Low-Resolution (LR) ones. This task has become one of the main problems in image and video processing because of the increasing demand for such methods from the industry. Due to the growing popularity of high-definition display devices, such as High-definition television (HDTV) and Ultra-high-definition television (UHDTV), there is an avid demand for HR videos. However, most of the contents (especially, older videos) have been obtained at much lower resolution. Therefore, there is a high demand for methods able to transfer LR videos into HR ones so that they can be displayed on HR TV screens, void of artifacts and noise.

In the problem of image Super-Resolution (SR), the high-to-low image formation model can be written as:

$$y = D(x \otimes k) + \epsilon, \quad (1)$$

This work was supported in part by the Sony 2016 Research Award Program Research Project. The work of SLT and RM was supported by the the Spanish Ministry of Economy and Competitiveness through project DPI2016-77869-C2-2-R and the Visiting Scholar program at the University of Granada. SLT received financial support through the Spanish FPU program. S. Lopez-Tapia and R. Molina are with the Computer Science and Artificial Intelligence Department, Universidad de Granada, Spain. A. Lucas and A.K. Katsaggelos are with the Dept. of Electrical Engineering and Computer Science, Northwestern University, Evanston, IL, USA. **This work has been submitted to the IEEE for possible publication. Copyright may be transferred without notice, after which this version may no longer be accessible.**

where $y$ is the LR image, $x$ is the HR image, $\epsilon$ is the noise, $x \otimes k$ represents the convolution of $x$ with the blur kernel $k$ and $D$ is a downsampling operator (usually chosen to be bicubic downsampling). In the case of Video Super-Resolution (VSR), $y$, $x$, and $\epsilon$ are indexed by a time index $t$ and additionally $\mathbf{y}_t$ is used to refer to the $2l+1$ LR frames in a time window around the HR center frame $x_t$, that is, $\mathbf{y}_t = \{y_{t-l},\ldots,y_t,\ldots,y_{t+l}\}$. Due to the strongly ill-posed nature of the SR problem, the recovery of the original HR image or video sequence is a very difficult task.

Current SR methods can be divided into two broad categories: model-based and learning-based algorithms. Model-based approaches explicitly define and use the process (blurring, sub-sampling and noise adding) by which LR image is obtained from the HR image or video sequence [1], [2], [3], [4]. With this explicit modeling, one can invert the SR model to obtain an estimate of the reconstructed HR frame. These methods rely on careful regularization to deal with the strong ill-posed nature of this problem. To enforce image-specific features into the estimated HR, signal priors are used, such as those controlling the smoothness or the total variation of the reconstructed image [1], [2], [3].

On the other hand, learning-based algorithms do not explicitly make use of the image formation model and use instead a large training database of HR and LR image/sequence pairs to learn to solve the SR problem. Recently, methods based on Convolutional Neural Networks (CNNs) have been proposed for SR and VSR, significantly surpassing classic learning-based and model-based methods. These methods try to find a function f(·) such that $x = $ f$(y)$ (or $x = $ f$(\mathbf{y})$ for VSR), which solves the mapping from LR images (or video sequences) to HR ones.

Although learning-based algorithms that use CNNs usually perform better than classic learning-based and model-based methods, they suffer from important limitations. The traditional approach to train CNNs for VSR is to first artificially synthesize a dataset with LR and HR pairs. The LR sequences at test time are assumed to have been subjected to the same degradation that was used during the training of the network. In other words, current CNN based methods are not robust to mismatch between training and testing degradation models, and so their performance greatly deteriorates [5] in that case. Since most of these methods are trained with the bicubic downsampling operator only, they cannot cope with changes in the degradation, which, in practice, significantly jeopardizes their application.

An additional problem with current CNN-based approaches to SR and VSR problem is that they are trained using the Mean-Squared-Error (MSE) cost function between the estimated SR frame and the HR frame. Numerous works in the literature (e.g., [6], [7], [8], [9]) have shown that while the MSE-based approach provides reasonable SR solutions, its conservative nature does not fully exploit the potential of Deep Neural Networks (DNNs) and produces blurry images. As an alternative to the MSE cost function, recent CNN-based SR methods use (during training) features learned by pre-trained discriminative networks and compute the $l_2$ distance between estimated and ground truth HR features. Using such feature-based losses in addition to the MSE loss has been proven to boost the quality of the super-resolved images [10]. Unfortunately, this approach introduces high-frequency artifacts. The use of Generative Adversarial Networks (GANs) [11] was also proposed as a mechanism to increase the perceptual quality of the estimated images trying to avoid again the smoothing introduced by the MSE (e.g., [12], [7], [13]).

In this work, we propose a new model that adapts the approximation proposed in [14] to Multiple-Degradation Video Super-Resolution (MDVSR). It uses the pseudo-inverse image formation model not only in the image formation model (as proposed in [14]), but also as an input to the network. Our experiments show that this model trained with MSE outperforms by far current state-of-the-art methods for bicubic degradation in terms of PSNR and SSIM metrics and it is significantly more robust to multiple degradations than current approaches. To further increase the sharpness of the resulting frames, we propose the use of a new loss function that combines adversarial and feature losses with a spatial smoothness constraint. This new loss allows for a significant increase in the perceptual quality of the estimated frames without producing the high-frequency artifacts typically observed with the use of GANs.

For all the described models, the use of an appropriate dataset is of paramount importance. While many of the current VSR learning-based models are trained using the Myanmar dataset, this dataset has limited variation of both scene types and motions. In this work we show that GAN-based VSR models significantly benefit from training with a dataset with more diverse scenes and motions. We obtain a significant increase in perceptual quality by training our best performing model on a dataset created from a subset of videos from the YouTube-8M dataset [15].

The rest of the paper is organized as follows. We provide a brief review of the current literature for learning-based VSR in Section II. In Section III, we present our baseline VSR model. In this section we detail the architecture used for our model. By additionally introducing our new spatial smoothness loss we obtain our proposed model trained to maximized perceptual quality. The training procedure, the new dataset, and experiments are described in detail in Section IV. In this section we also evaluate the performance of the proposed models by comparing them to current state-of-the-art VSR approaches for scale factors 2, 3, 4 and different degradations. Our quantitative and qualitative results show that our proposed perceptual model sharpens the frames to a much greater extent than current VSR state-of-the-art DNNs without the introduction of artifacts. In addition, we show in this section that our resulting model is far more robust to variations in the degradation model compared with the current state-of-the-art model. Finally, conclusions are drawn in Section V.

## II. RELATED WORK

In recent years, different VSR CNN-based models have been proposed in the literature. Liao et al. [16] utilize a two-step procedure where an ensemble of SR solutions is first obtained through the use of an analytic approach. This ensemble then becomes the input to a CNN that calculates the final SR solution. Kappeler et al. [17] use a three layer CNN to learn a direct mapping between the bicubically interpolated and motion compensated LR frames in $\mathbf{y}_t$ and the corresponding HR central frame $x_t$. Other works have applied Recurrent Neural Networks (RNNs) to VSR. For example, in [18] the authors use a bidirectional RNN to learn from past and future frames in the input LR sequence. Although RNNs have the advantage of exploiting more effectively the temporal dependencies between frames, the challenges and difficulties associated to their training has led to CNN being the favored DNN for VSR. Li and Wang [19] exploit the benefits of residual learning with CNNs in VSR by predicting only the residuals between the high-frequency and low-frequency frame. Caballero et al. [20] jointly train a spatial transformer network and a CNN to warp video frames, they then benefit from sub-pixel information, avoiding the use of motion compensation (MC). Similarly, Makansi et al. [21] and Tao et al. [22] found that jointly performing upsampling and MC increases the performance of the VSR model.

All these previous methods use the MSE loss as the cost function during the training phase. This is the most common practice in the literature for CNN-based models. However, the use of this loss during training causes the estimated frames to be blurred. In an attempt to solve this problem, recent works have used feature-based losses as additional cost functions, see [6]. This approach has significantly improved the sharpness and the perceptual quality of the estimations. Ledig et al. [7] proposed a combination of a GAN and feature losses for training, leading to the generation of images with superior photorealistic quality. In [10], Lucas et al. proposed an adaptation of this approach to VSR. They introduced a new loss based on a combination of perceptual features and the use of the GAN formulation. The model has led to a new state-of-the-art for VSR in terms of perceptual quality. To improve the quality of the predicted images, Wang et al. [8] train a GAN for image SR conditioning the output of the network using semantic information extracted by a segmentation CNN. In [9] the authors propose the Residual-in-Residual Dense Block and use it to construct a very deep network that is trained for image SR using perceptual losses.

As previously stated in Section I, SR and VSR methods can be classified into two groups: model based and learning based. However, recently new methods that blend the two approaches have emerged. In [23], Zhang et al. use the Alternating Direction Method of Multipliers (ADMM) for image recovering problems with known linear degradation models, such as image deconvolution, blind image deconvolution, and SR. ADMM methods split the recovery problem into two subproblems: a regularized recovery one (subproblem A) and a denoising one (subproblem B). The authors of [23] propose to combine learning and analytical approaches by using a CNN for the denoising problem. This allows them to use the same network for multiple ill posed inverse imaging problems. At the same time, some works have been proposed to increase the performance and the flexibility of SR learning-based models by taking into account the image formation model when training their CNNs. More specifically, Sonderby et al. [14] proposed a new approach which estimates and explicitly uses the image formation model to learn the solution modeled by the network. The blurring and downsampling process to obtain LR frames from HR ones is estimated and the Maximum a Posteriori (MAP) HR image estimation procedure is approximated with the use of a GAN. We improve over this approach by generalizing it to multiple degradation operators for the VSR. To enforce robustness to multiple degradations in the case of single image SR, Zhang et al. [24] propose to input to their CNN not only the LR image but also the Principal Component Analysis (PCA) representation of the blur kernel used in the degradation process. We adopt a similar approach in our framework, as detailed in the next section. Preliminary results of our approach can be found in [25].

## III. Model description

In this section, we first introduce the problem of VSR with multiple degradations and explain how we can adapt the Amortised MAP approximation in [14] to solve it. We then introduce a new perceptual loss based on the loss proposed in [10] and incorporate a spatial smoothness constraint to avoid the high frequency artifacts produced by perceptual losses while retaining the increase in overall sharpness of the predicted frames. Finally, we describe our proposed new architecture based on the VSRResNet architecture originally introduced in [10]. This architectural modification aims to improve the robustness to multiple degradations, that is, not restricting high performance to a single fixed $D$ and $k$ in Eq. 1.

As previously stated in Section I, we use $x$ to denote a HR frame in a video sequence and $y$ its corresponding observed LR frame. Furthermore, we use $\mathbf{y}$ to refer to the LR frames in a time window around the HR center frame $x$, $\mathbf{y}$ contains $2l+1$ frames and we use $l=2$ in the experiments.

The process of obtaining a LR image from the HR one is usually modeled using Eq. 1. In this paper, we assume that the image formation noise is negligible ($\epsilon = 0$) and absorbed by the downsampling process. Also, following previous works in the literature like [24], we assume that $D$ represents bicubic downsampling and the blur $k$ is known and has the form of an isotropic Gaussian kernel. Assuming that the blur is known is not a major constraint since it can be reasonably approximated using any of the techniques proposed in [26], [27], [28]. These techniques, which are used in blind SR, aim at estimating $k$ rather than increasing the quality of the SR solution, therefore we can think of these approaches as complementary to our method. In addition our experiments show that the estimated kernels are in practice accurate enough to ultimately produce frames of high quality. Although more complex blurs, like motion blur, can also be considered, our downsampling and Gaussian blur model is frequently assumed to be a good representation of the high to low degradation process [24]. Notice that, although $k$ and $x$ could be estimated at the same time, this does not work well in practice [24]. A blind approach without a specially designed architecture has poor generalization ability and particularly aggravates the pixel average problem [7], which goes against our objectives of obtaining a model robust to multiple degradation operators capable of producing frames of high quality. We also assume that all the frames in the time window are degraded with the same operator. Since this operator depends mostly on the camera used, it is unreasonable to assume that these conditions will change drastically from one frame to another. In summary, we assume that $D$ (bicubic downsampling) and $k$ (Gaussian blur) in Eq. 1 are known (or previously estimated) and that they are constant for all frames in $\mathbf{y}$. Notice that assuming that the Gaussian blur is known is not the same as to assume that it is the same for all the video sequences. The use of this image forward model leads to a more challenging VSR problem than when only bicubic downsampling is considered, which is the modelling used in most previous works on VSR, see [10], [16], [19], [21], [22], [29], [17].

Let us now examine how we can approach the VSR with multiple degradations problem. Most current SR methods solve the problem by learning a function $f_\theta(.)$, which maps a low resolution image to the high resolution space, using training data pairs $x$ and $\mathbf{y}$ and optimizing the parameters $\theta$ using gradient descent over a *Mean Square Error* function. This model embeds the estimation of the downsampling process in the function $f_\theta(\mathbf{y})$. We argue here that to obtain a SR network capable of dealing with multiple degradations, it is necessary to separate its learning process from the degradation as much as possible. To achieve this, given $D$ and $k$, we define $A = Dk$ as proposed in [14] to our MDVSR problem and consider the function

$$g_\theta(\mathbf{y}) = (I - A^+ A)f_\theta(\mathbf{y}) + A^+ y, \qquad (2)$$

where $A^+$ denotes the Moore-Penrose pseudoinverse of the degradation $A$. Since $AA^+A = A$ and $A^+AA^+ = A^+$, and because the rows of $A$ are independent $AA^+ = I$, we have

$$Ag_\theta(\mathbf{y}) = A(I - A^+A)f_\theta(\mathbf{y}) + AA^+y = y \qquad (3)$$

The resulting $g_\theta(\mathbf{y})$ is a HR image which satisfies Eq. 1 when $\epsilon = 0$. With this formulation, the learning of the network is made easier by learning a residual only. This has been exploited in other works of SR where the networks learn to predict a residual over an initial estimation, usually chosen to be the bicubic interpolation [30]. However, these other approaches are not well suited for the Multiple Degrations setting, since the quality of the initial prediction may vary significantly from one degradation to another, showing different kind of artifacts. Fig. 1 illustrates this problem for bicubic interpolation.

Notice that in order to use our approach, the estimation of the $A^+$ operator prior to training is required. In [14], this operator is modeled using a convolution operation followed by a subpixel shuffle layer [31]. The unknown parameters $w$ are estimated by minimizing, via stochastic gradient descent, the loss function, that is,

$$\hat{\omega} = \arg\min_{\omega} \mathbb{E}_x \| Ax - AA_{\omega}^+(Ax) \|_2^2 \\ + \mathbb{E}_y \| A_{\omega}^+(y) - A_{\omega}^+(AA_{\omega}^+(y)) \|_2^2, \quad (4)$$

where $A_{\omega}^+$ denotes the pseudo-inverse with $\omega$ network parameters.

An obvious disadvantage of this approach is that one needs to learn a specific $A_{\hat{w}}^+$ for each $A$. In order to have a single network robust to multiple $A$ operators, we have implemented a network that, for any given $A$, predicts the corresponding $\hat{\omega}$ of its pseudoinverse. We choose this network to be composed of three hidden layers with 512, 1024 and 512 hidden units. The input to this network is the PCA representation of the kernel $k$ of $A$. The network is trained to predict the unknown $\hat{\omega}$ which solves Eq. 4 for a given $A$. Our experiments found that the performance obtained by this efficient approach was equivalent to calculating $\hat{\omega}$ for each $A$ by individually solving Eq. 4. for each operator.

Let us now see how $g_{\theta}(\mathbf{y})$ is obtained using an enhanced formulation of a GAN model with increased perceptual quality.

### A. A GAN model with increased perceptual quality

Taking into account that the transformation $g_{\theta}(\mathbf{y})$ defines (from the distribution on $\mathbf{y}$) a probability distribution function $q_{\theta}(.)$ on the set of HR images, the Kullback-Leibler divergence between $q_{\theta}(.)$ and the distribution of real images $p_X(.)$

$$\mathrm{KL}(q_{\theta} \| p_X) = \int q_{\theta}(x) \log \frac{q_{\theta}(x)}{p_X(x)} dx \quad (5)$$

is minimized using a GAN approach. This model has a maximum a posteriori approximation interpretation (see [14] for the details).

Together with the generative network, $g_{\theta}(\mathbf{y})$, we learn a discriminative one, $d_{\phi}(x)$, using the following two functions on $\phi$ and $\theta$

$$\mathrm{L}(\phi; \theta) = -\mathbb{E}_{x \sim X} \left[ \log d_{\phi}(x) \right] - \mathbb{E}_{\mathbf{y} \sim Y} \left[ \log(1 - d_{\phi}(g_{\theta}(\mathbf{y}))) \right] \quad (6)$$

$$\mathrm{L}(\theta; \phi) = -\mathbb{E}_{\mathbf{y} \sim Y} \left[ \log \frac{d_{\phi}(g_{\theta}(\mathbf{y}))}{1 - d_{\phi}(g_{\theta}(\mathbf{y}))} \right]. \quad (7)$$

where X and Y are the distribution of the HR and LR images respectively. Iteratively, the algorithm updates $\phi$ by lowering $\mathrm{L}(\phi; \theta)$ while keeping $\theta$ fixed, and updates $\theta$ by lowering $\mathrm{L}(\theta; \phi)$ while keeping $\phi$ fixed.

Notice that because the difference between $\mathbb{H}[q_G, p_X]$ (cross-entropy) and $\mathrm{KL}[q_G | p_X]$ is $\mathbb{H}[q_G]$, this approximation is expected to lead to solutions with higher entropy and so produce more diverse frames, see [14]. As can be seen in Fig. 2, this solution leads to good results for factor 2, however, it becomes unstable for larger scale factors such as 3 and 4, where the GAN failed to converge. This is most likely caused by the discriminator's ability to easily distinguish between real and generated frames (furthermore, notice that the generator has to produce 16 pixels in the HR frame for each pixel in the input LR frame, which is a considerably more challenging task).

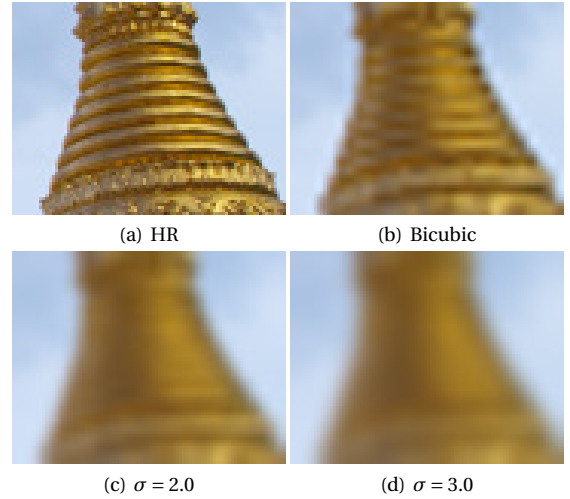

Fig. 1: Example of artifacts introduced by the bicubic downsampling interpolation for a scaling factor of 3. (a) shows the original image, (b) corresponds to bicubically downsampling the image in (a), (c) and (d) show bicubically downsampled images which have previously been blurred with $\sigma = 2$ and $\sigma = 3$ Gaussian kernels. The downsampled images have been enlarged to the size of the original one using bicubic upsampling.

In order to regularize [32] the training of our GAN network, we follow the approach described in [10] and use the Charbonnier loss between two images $u$ and $v$ (in a given space) defined as

$$\gamma(u, v) = \sum_k \sum_i \sum_j \sqrt{(u_{k,i,j} - v_{k,i,j})^2 + \epsilon^2}. \quad (8)$$

The Charbonnier loss is calculated in both pixel and feature spaces. We define our feature space to correspond to the activations provided by a CNN trained for discriminative tasks. For our model we used the third and fourth convolution convolutional layers of VGG-16 [33] (denoted as VGG(.)).

The combined loss proposed by [10] becomes:

$$\mathrm{L}_{\text{total}}(\theta; \phi) = \alpha \sum_{(x, \mathbf{y}) \in T} \gamma(\mathrm{VGG}(x), \mathrm{VGG}(g_{\theta}(\mathbf{y}))) \\ + \beta \mathbb{E}_{\mathbf{y}} \left[ \log \frac{1 - d_{\phi}(g_{\theta}(\mathbf{y}))}{d_{\phi}(g_{\theta}(\mathbf{y}))} \right] + (1 - \alpha - \beta) \sum_{(x, \mathbf{y}) \in T} \gamma(x, g_{\theta}(\mathbf{y})) \quad (9)$$

where $\alpha, \beta > 0$, $\alpha + \beta < 1$ and $T$ is the dataset formed by paris of LR sequences $\mathbf{y}$ and HR images $x$.

While this loss successfully produces frames of previously unseen sharpness in VSR, it also introduces high frequency artifacts, especially in smooth areas of the image. They are easily detectable and unpleasant to the human eye (see Fig. 3 for examples of such artifacts). While increasing the weight of the pixel-content loss ($\sum_{(x,\mathbf{y})\in T}\gamma(x,g_\theta(\mathbf{y}))$) significantly reduces these artifacts, it also smoothes the frame. Because these artifacts are more prominent on smooth regions of the frame, we propose to use a weight matrix M($x$) that assigns more weight to the pixel-content loss in the smooth areas of the real HR frame $x$ during training. With the incorporation of this spatial smoothness constraint, the generator will be more penalized when generating unwanted noise in smooth regions of the frame. We compute this weight matrix as M($x$) = 1 − S($x$). where S($x$) is the Sobel operator applied to image $x$. Therefore, the new proposed loss becomes:

$$L_{\text{totalsmooth}}(\theta;\phi) = \alpha \sum_{(x,\mathbf{y})\in T} \gamma(\text{VGG}(x), \text{VGG}(g_\theta(\mathbf{y})))$$
$$+ \beta[\mathbb{E}_\mathbf{y}[\log \frac{1-d_\phi(g_\theta(\mathbf{y}))}{d_\phi(g_\theta(\mathbf{y}))}]] + (1-\alpha-\beta) \sum_{(x,\mathbf{y})\in T} M(x) \odot \gamma(x, g_\theta(\mathbf{y}))$$
(10)

where $\alpha, \beta > 0$ and $\alpha + \beta < 1$ and $\odot$ denotes element-wise multiplication.

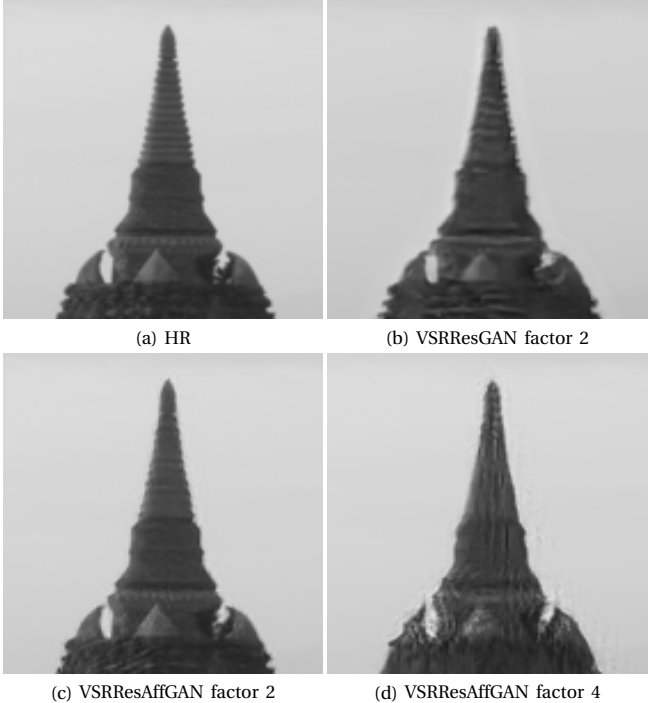

(a) HR  (b) VSRResGAN factor 2
(c) VSRResAffGAN factor 2  (d) VSRResAffGAN factor 4

Fig. 2: Qualitative comparison between our GAN model (VSRResAffGAN) and [10] (VSRResGAN) when only adversarial loss is used. We can see how our model is able to recover the frame for factor 2 while [10] fails producing a lot of artifacts and a blur frame.

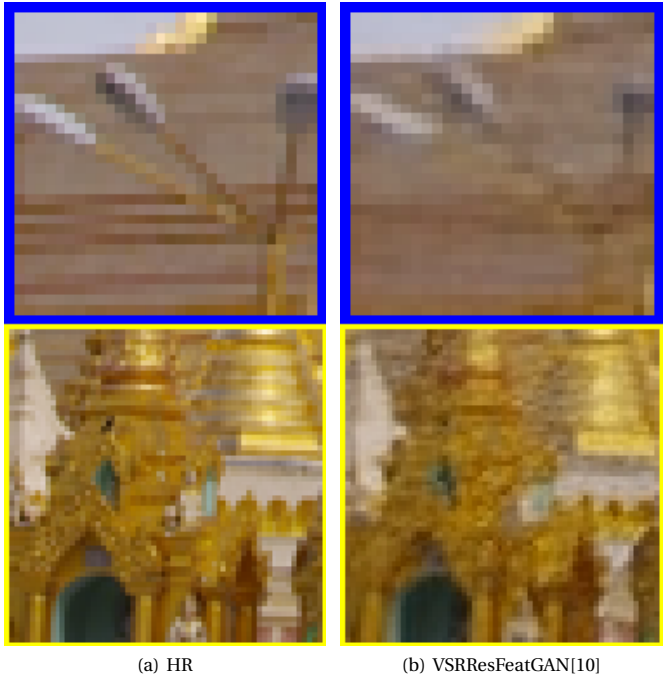

(a) HR  (b) VSRResFeatGAN[10]

Fig. 3: Qualitative results for factor 3 of VSRResFeatGAN[10]. Notice the high frequency artifacts on smooth areas of the image.

In the next section, we describe in detail the CNN architecture used to approximate $f_\theta(\cdot)$.

### B. Architecture

To implement $f_\theta(\cdot)$, from which we will obtain $g_\theta(y)$ using Eq. 2 which will then be used in Eq. 10, we adapt the VSRResNet model introduced in [10]. The authors of [10] found that the VSRResNet architecture results in state-of-the-art performance on the VideoSet4 dataset [34], the test dataset commonly used for evaluating VSR models. The VSRResNet model corresponds to a deep residual CNN that consists of 3 3×3 convolutional layers each followed by a ReLU activation, 15 Residuals Blocks with no batch normalization and a final 3×3 convolutional layer. Padding is used at each convolution step in order to keep the spatial extent of the feature maps fixed across the network.

We note here that using as input the bicubically upsampled frames as in [10] is not well suited for the multiple degradation setting established in our work. Bicubic upsampling over-smoothes the images and introduces artifacts, making the learning process more difficult. Furthermore, as shown in Fig. 1, these artifacts differ from one degradation operator to another. Instead, we decided to input the LR video sequence to the network and use the sub-pixel shuffle layer introduced in [31] in the network architecture to learn the upscaling operation. This avoids the previously mentioned problem and increases training and inference speed.

The architecture defined above still suffers from a major problem: the parameters of the network $\theta$ depend on the choice of $A$. Although we ease the training procedure by only predicting the residual using $(I - A_{\hat{\omega}}^+ A)f_\theta(\mathbf{y})$, the network parameters are dependent on $A_{\hat{\omega}}^+ y$. In the MDVSR setting, it is necessary for the network parameters to be independent of $A$. This will allow any input video sequence to be provided to the trained network at test time. To this end, we modify the network architecture such that knowledge of $A$ is provided. This will allow the network parameters to be learned for all $A$s and to adapt to any given $A$ at test time. More specifically, we encode $A_{\hat{\omega}}^+ y$ using a network $e_\psi(\cdot)$ ($e_\psi$ in Fig. 4) and feed this compressed representation of $A^+_{\hat{\omega}} y$ to the CNN. This encoder network $e_\psi(\cdot)$ seeks to extract the relevant and significant information from the degradation operator to guide the SR process. Notice that other approaches, such as utilizing the Principal Components of $k$, see [24], may be considered. However, we argue that using an encoder network is more appropriate as more useful information can be extracted than by using a PCA representation. Our encoder $e_\psi(A^+_{\hat{\omega}} y)$ consists of three convolutions of 3 × 3 and 32 filters with zero-padding, followed by the ReLU activation. Our best results were obtained by incorporating $e_\psi(A^+_{\hat{\omega}} y)$ by concatenating the resulting feature maps before the first residual block of the VSRResNet architecture, see Fig. 4 for details. To ensure that the spatial size matches that of **y** we use a convolution stride equal to the scaling factor used for training. We jointly train the encoder $e_\psi(\cdot)$ and the super-resolving $f_\theta(\cdot)$ network.

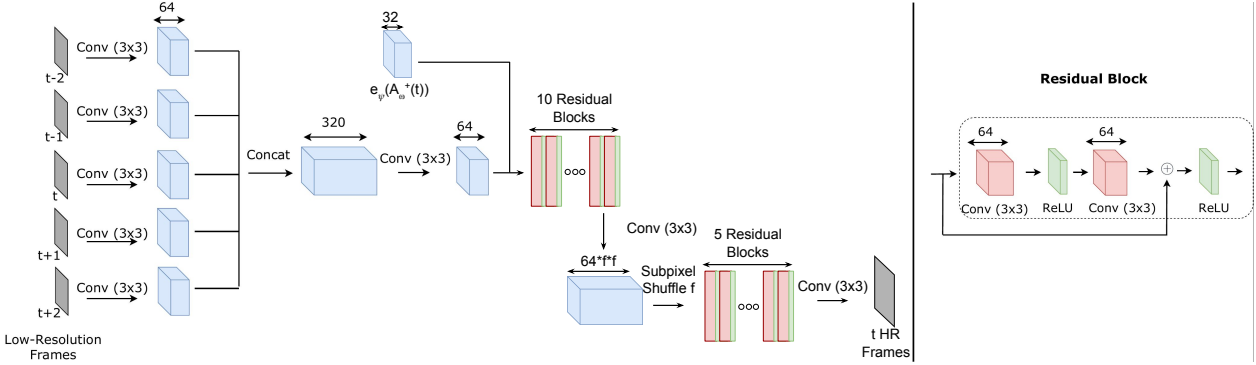

Fig. 4: The MD-AVSR architecture based on VSRResNet[10]. The network consists of a series of convolution operations with 64 kernels of size 3 × 3, applied on each input frame. The resulting feature maps are then concatenated together to obtain 320 feature maps. This is followed by two convolution operations and 15 residual blocks. Each residual block consists of two convolutional operations with 64 kernels of size 3 × 3, each followed by a ReLU layer. Following the definition of a residual block, the inputted feature maps are added to the output feature maps to obtain the final output of the residual block. Before the 10th Residual block, we use upscale the feature maps by a factor $f$ using a sub-pixel shuffle layer [31].

## IV. Experimental Results

In this section we will show that our proposed approach significantly outperforms current state-of-the-art models for bicubic degradation in addition to being robust to variation on the Gaussian blur deviation.

We use two datasets to train our models: The training sequences from the Myanmar dataset and a second one extracted from a subset of the YouTube-8M dataset, which will be used to refine our best performing model.

Let us start by using the Myanmar training sequences. The training dataset is formed by $10^6$ patches of size 48 × 48 pixels. Patches with variance less than 0.0035 were determined as being uninformative and were hence removed from the dataset. For each HR patch at time $t$, we obtain the corresponding LR sequence of patches at time $t-2$, $t-1$, $t$, $t+1$, and $t+2$. These LR patches are obtained following Eq. 1, i.e., by first blurring the HR frames with a Gaussian kernel and then downsampling the images using bicubic downsampling.

To determine the quality of the new proposed architecture, we compare it to other proposed architectures for VSR. For a fair comparison, we will use neither the GAN framework nor the perceptual losses. Instead, we train using the MSE loss ($\mathbb{E}_{x,\mathbf{y}}[\|x - g_\theta(\mathbf{y})\|^2]$). We use the Adam optimizer [35] for 100 epochs. The learning rate was set to $10^{-3}$ for the first 50 epochs and then divided by 10 at the 50th and 75th epochs. The weight decay parameter was set to $10^{-5}$ for all our trained models. For the first set of experiments, we fix the degradation operator to correspond to the bicubic downsampling of factor 3 (no Gaussian blur). Because we only use one degradation for these experiments, we temporarily remove the use of $e_\psi(A^+_{\hat{\omega}} y))$ from our architecture and instead use as input the LR **y** only.

We name the model that uses $g_\theta(\cdot)$, i.e. the model that utilizes an affine projection, Affine VSR (AVSR) and the model that uses $f_\theta(\cdot)$ No Affine VSR (NoAVSR) (it minimizes $\mathbb{E}_{x,\mathbf{y}}\|x - f_\theta(\mathbf{y})\|^2$). To determine the contribution of the sub-pixel shuffle layer, we also train an architecture similar to AVSR but using bicubic upsampling at the input instead of using the subpixel shuffle layer. We name this model Bicubic AVSR (B-AVSR). Notice that the VSRResNet model introduced in [10] corresponds to our NoAVSR model with a bicubically interpolated input instead of a sub-pixel layer used in NoAVSR.

Table I.a shows the results of our quantitative comparisons of these models for multiple degradations and upscaling factor 3 on the test sequences of the Myanmar dataset. From this table it is clear that the proposed model AVSR significantly outperforms the other models. However, as expected, all of the models perform worse for degradations different from bicubic downsampling, as these were trained to expect the bicubic downsampling operator only. Also, notice how using the subpixel shuffle layer does not help non-affine architectures like NoAVSR. Notice also the increased performance when $g_\theta(\cdot)$ (the affine projection) is used as it is demonstrated by the increase in performance from B-AVSR to AVSR presented in Table I.a. In summary, this experiment indicates that explicitly using the affine transformation when training improves the quality of the reconstruction, it also shows that it is more convenient to use as input the LR sequence instead of the bicubically interpolated one. Unfortunately, it also shows that the AVSR is not robust to the use, during testing, of images which have been degraded with Gaussian blur when trained only with bicubic downsampling. Notice however that our AVSR is the best performing one even for this case.

To determine the performance of our proposed model with multiple degradations, we train our model ($g_\theta(\cdot)$) which uses the encoded information $e_\psi(A_{\tilde{\omega}}^+ y))$ as input to our network architecture, with multiple degradations and factors. We call this model MD-AVSR. Notice that we are still using just MSE to train these models, i.e., no GAN formulation. We also trained the Blind MD-AVSR (B-MD-AVSR) model, which uses bicubic upsampling at the input. The degradations considered here are a combination of Gaussian blurs with different kernels $k$ and bicubic downsampling. We generated random Gaussian kernels with $\sigma$ using a step of 0.1 in the range [0.2, 2.0] for factor 2, [0.2, 3.0] for factor 3 and [0.2, 4.0] for factor 4. After that, the HR video sequences are blurred with these kernels and bicubic downsampling is applied to them to generate the LR samples.

To provide a comparison with current state-of-the-art methods we also adapt the SRMDNF model proposed in [24]. It uses PCA($k$) and incorporates this information into an approach that uses $f_\theta(\cdot)$ and not $g_\theta(\cdot)$. We have experimentally determined that the optimal place to add this information is before the first residual block. We trained this network as we did with MD-AVSR. We call this model VSRMDNF, see [36], we also compared with the SRMDNF method in [24].

|  | Bicubic PSNR/SSIM | $\sigma$ = 1.0 PSNR/SSIM | $\sigma$ = 2.0 PSNR/SSIM | $\sigma$ = 3.0 PSNR/SSIM |
|---|---|---|---|---|
| VSRResNet | 35.97/0.9481 | 33.39/0.9210 | 29.09/0.8365 | 27.18/0.7680 |
| NoAVSR | 35.92/0.9474 | 33.39/0.9156 | 29.16/0.8362 | 27.18/0.7678 |
| B-AVSR | 36.09/0.9487 | 33.41/0.9214 | 29.13/0.8374 | 27.20/0.7685 |
| **AVSR** | **36.35/0.9522** | **33.52/0.9279** | **29.23/0.8454** | **27.42/0.7836** |

(a) Models trained only used bicubic degradation.

|  | Bicubic PSNR/SSIM | $\sigma$ = 1.0 PSNR/SSIM | $\sigma$ = 2.0 PSNR/SSIM | $\sigma$ = 3.0 PSNR/SSIM |
|---|---|---|---|---|
| IRCNN[36] | 34.41/0.8937 | 34.44/0.8937 | 33.58/0.8937 | 29.92/0.8937 |
| SRMDNF[24] | 35.08/0.9299 | 35.14/0.9298 | 34.78/0.9224 | 33.20/0.8937 |
| B-MD-AVSR | 34.97/0.9330 | 34.59/0.9275 | 34.27/0.9200 | 34.62/0.9270 |
| VSRMDNF | 35.95/0.9471 | 35.82/0.9439 | 35.28/0.9365 | 35.01/0.9319 |
| **MD-AVSR** | **36.52/0.9525** | **36.27/0.9494** | **35.60/0.9406** | **35.22/0.9352** |

(b) Models trained with multiple degradations.

TABLE I: Comparison of the proposed and state-of-the-art models for Myanmar dataset test sequences for factor 3. $\sigma$ refers to the Gaussian blur deviation used.

Table I.b shows an experimental comparison with current state-of-the-art models for multiple degradations and factors 2, 3, and 4 on the VidSet4 video sequences [34].

|  | Factor 2 PSNR/SSIM/ | Factor 3 PSNR/SSIM/ | Factor 4 PSNR/SSIM |
|---|---|---|---|
| VSRResNet | 31.87/0.9426 | 27.80/0.8571 | 25.51/0.753 |
| VSRResFeatGAN | 30.90/0.9241 | 26.53/0.8148 | 24.50/0.7023 |
| ERSGAN[9] | × | × | 22.98/0.6336 |
| **MD-AVSR** | **33.00/0.9496** | **28.31/0.8751** | **26.17/0.7895** |
| MD-AVSR-FG | 31.54/0.9309 | 26.73/0.8237 | 25.10/0.7414 |
| MD-AVSR-P | 31.82/0.9345 | 27.20/0.8383 | 25.26/0.7501 |
| MD-AVSR-PY8 | 31.81/0.9391 | 27.09/0.8412 | 25.18/0.7555 |
|  | Factor 2 PercepDist | Factor 3 PercepDist | Factor 4 PercepDist |
| VSRResNet | 0.0407 | 0.1209 | 0.1766 |
| VSRResFeatGAN | 0.0283 | 0.0668 | 0.1043 |
| ERSGAN[9] | × | × | 0.0993 |
| MD-AVSR | 0.0292 | 0.1081 | 0.1655 |
| MD-AVSR-FG | 0.0229 | 0.0588 | 0.0939 |
| MD-AVSR-P | 0.0216 | 0.0586 | 0.0927 |
| **MD-AVSR-PY8** | **0.0210** | **0.0571** | **0.0916** |

TABLE II: Comparison with state-of-the-art methods on the VidSet4 dataset on scale factors 2, 3, and 4. The first table uses PSNR and SSIM and the second table uses the Perceptual Distance as defined in [37]. A smaller Perceptual Distance implies better perceptual quality. Values for factors 2 and 3 for ERSGAN could not be calculated since the authors did not train the network for those.

We can see that the proposed MD-AVSR outperforms all the other models for all values of $\sigma$ considered. Notice that B-MD-AVSR suffers from a sharp decrease in performance. This indicates the need to use the degradation information if we intend to utilize the same network with multiple degradations. Notice also that MD-AVSR outperforms VSRMDNF in a similar manner as AVSR outperforms NoAVSR and VSRResNet. This indicates that the benefits of using the image formation model are carried over to the multiple degradation setting. It is important to observe in Table I that AVSR and MD-AVSR are the best performing methods when compared to similar approaches. Furthermore, for bicubic downsampling MD-AVSR slightly outperforms AVSR which is expected to deal with this degradation only. Notice also that although the performance of MD-AVSR slightly deteriorates when blur is introduced, it is much more robust to this degradation than its AVSR counterpart, in other words, MD-AVSR could be safely implemented in systems that are expected to deal with video sequences degraded by different acquisition models. A qualitative comparison of VSRResNet, AVSR, VSRMDNF and MD-AVSR can be seen in figures 5 and 6. This comparison reveals how our methods are more robust to the multiple degradations and produce HR images of higher quality.

Let us now test the GAN framework and analyze the contribution of the new loss in comparison to the one proposed in [10]. Using the loss proposed in [10] we fine-tuned (as before we utilize multiple degradations to generate the LR examples) the MD-AVSR model already trained with MSE with the loss proposed in [10] for 30 epochs. The learning rate and weight decay for the discriminator are set to $10^{-4}$ and $10^{-3}$, respectively while for the generator they are both set to $10^{-4}$. The learning rate of both Generator and Discriminator are divided by 10 after 15 epochs. The values of the hyper-parameters are: $\alpha = 0.998$ and $\beta = 0.001$. We call this model MD-AVSR-FG. We fine-tuned in the same fashion the MD-AVSR model with the new proposed loss and call it MD-AVSR-P. The values of the hyper-parameters are: $\alpha = 0.049$ and $\beta = 0.001$. Finally, to demonstrate the influence of the diversity of the training dataset for GAN-based VSR models, we train our MD-AVSR-P model on 1306844 blocks of size 48 × 48 from 5 consecutive images extracted from a subset of YouTube-8M dataset. This subset was constructed by randomly selecting a total of 4358 videos. We consider all categories except those corresponding to video games and cartoons since these categories do not provide an accurate representation of the natural scenes we are interested in recovering in VSR.

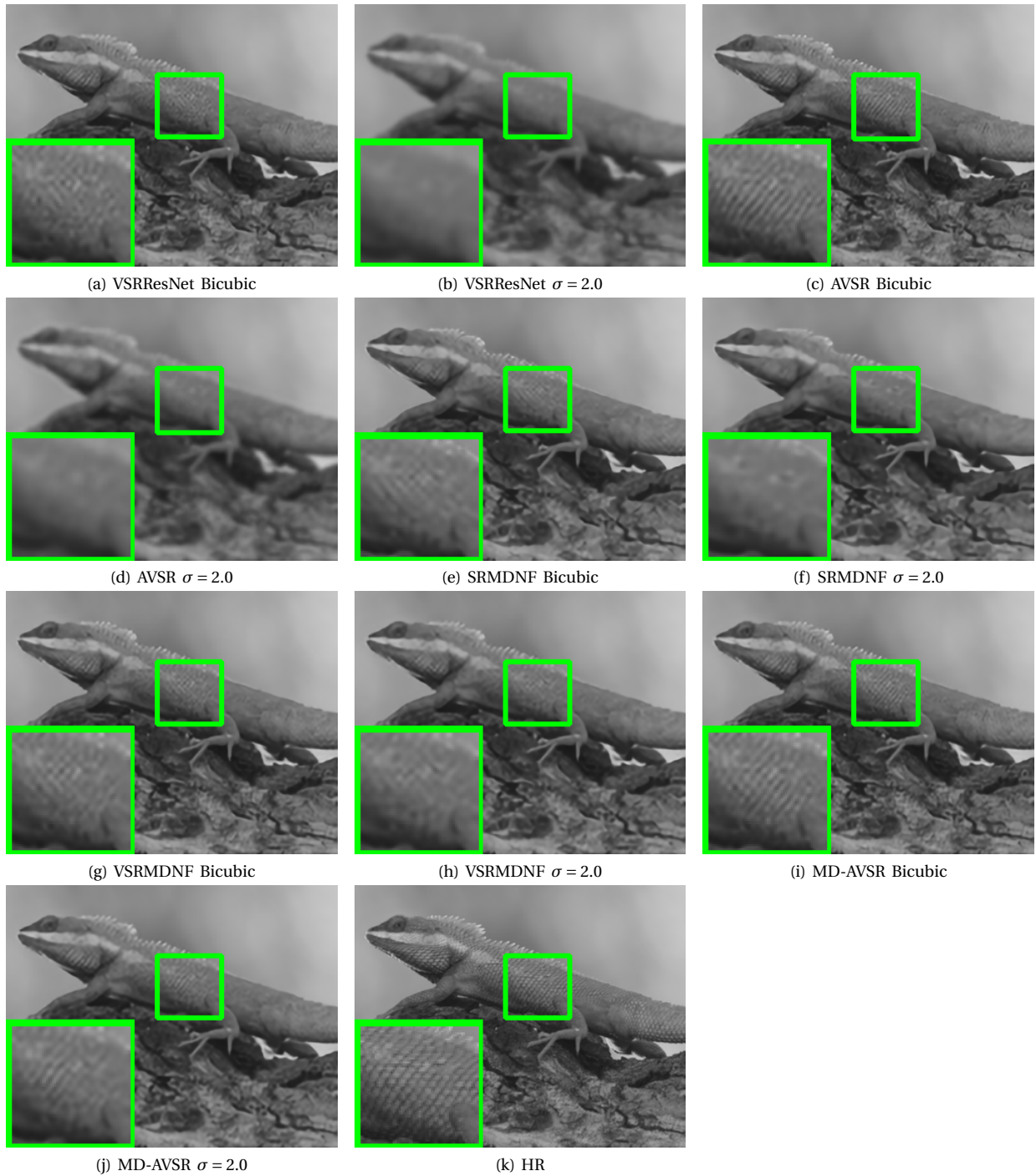

Fig. 5: Qualitative results of our VSR model compared to other state-of-the-art methods for factor 3 using bicubic downsampling and Gaussian blur with $\sigma = 2.0$ and bicubic downsampling. Notice how MD-AVSR recovers more details compared to the rest.

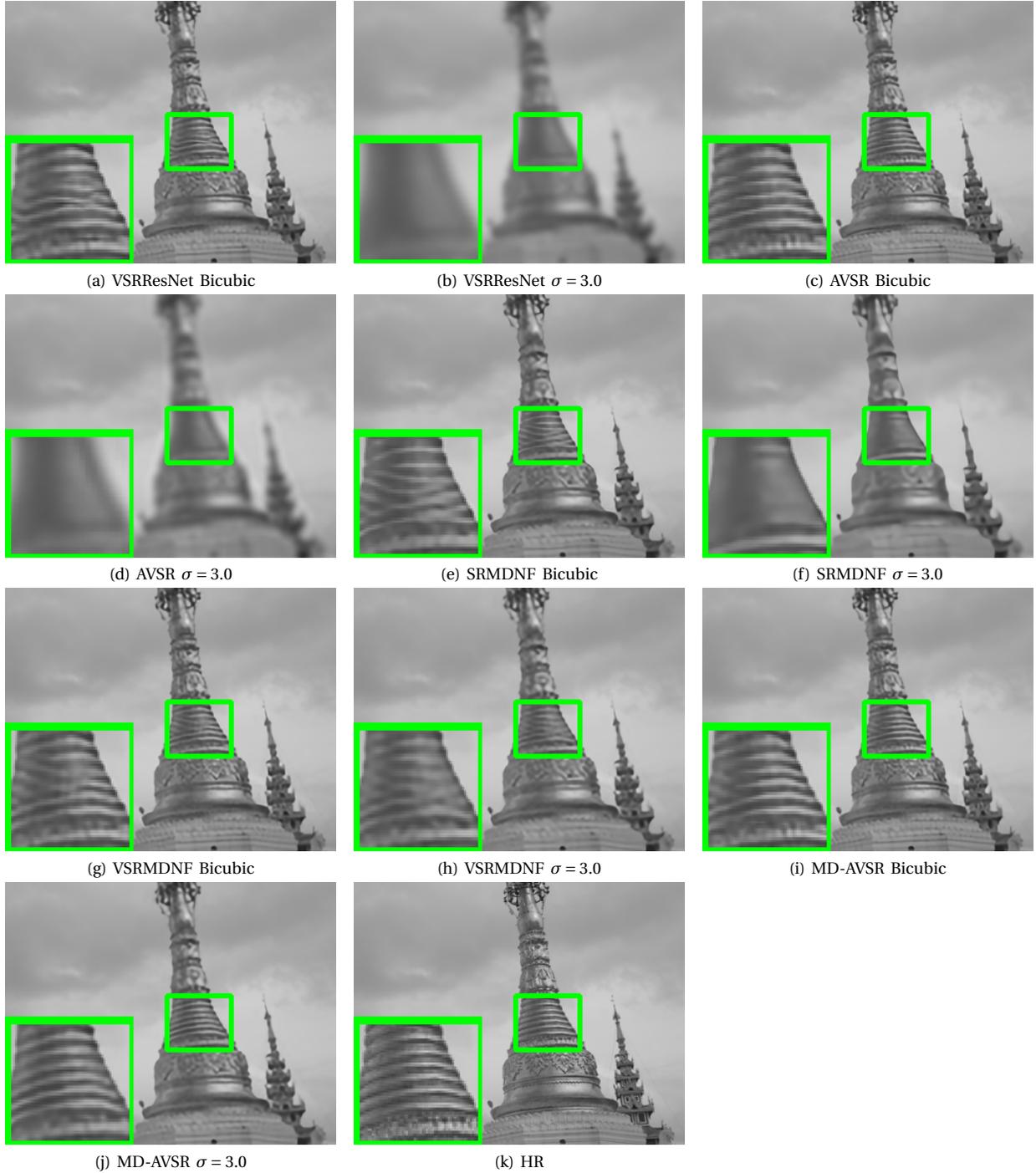

Fig. 6: Qualitative results of our VSR model compared to other state-of-the-art methods for factor 3 using bicubic downsampling and Gaussian blur with $\sigma = 3.0$ and bicubic downsampling. The pattern highlighted in the HR image is difficult to recover when only bicubic downsampling is used because it breaks the lines (see Fig. 1). In this case, our methods is able to almost fully recover the correct structure while the rest methods struggle. Furthermore, our method is able to fully recover the correct pattern when a strong Gaussian blur of $\sigma = 3.0$ is applied, while the other methods fail to do so.

We call this model MD-AVSR-PY8.

Table II contains a comparison of these models with current state-of-the-art methods in terms of PSNR, SSIM and Perceptual Distance [37] for the VidSet4. We can see how the proposed MD-AVSR-P outperforms all the models trained with perceptual losses for all the figures of merits when trained on Myanmar train sequences. Furthermore, a close examination of the generated frames and a compar-ison to the MD-AVSR-FG ones (see Fig. 7) shows that they are almost artifact-free. The model also shows a noticeable increase in sharpness compared to MD-AVSR. However, this increase in sharpness (reflected on the improvement on Perceptual Distance) comes with a decrease in PSNR and SSIM. Fig. 8 shows a qualitative comparison of VSRResFeatGAN and ERSGAN with MD-AVSR-P. We can see that the proposed model MD-AVSR-P outperforms current state-of-the-art methods. Notice that values for factors 2 and 3 for ERSGAN could not be calculated since the authors did provide weights for those.

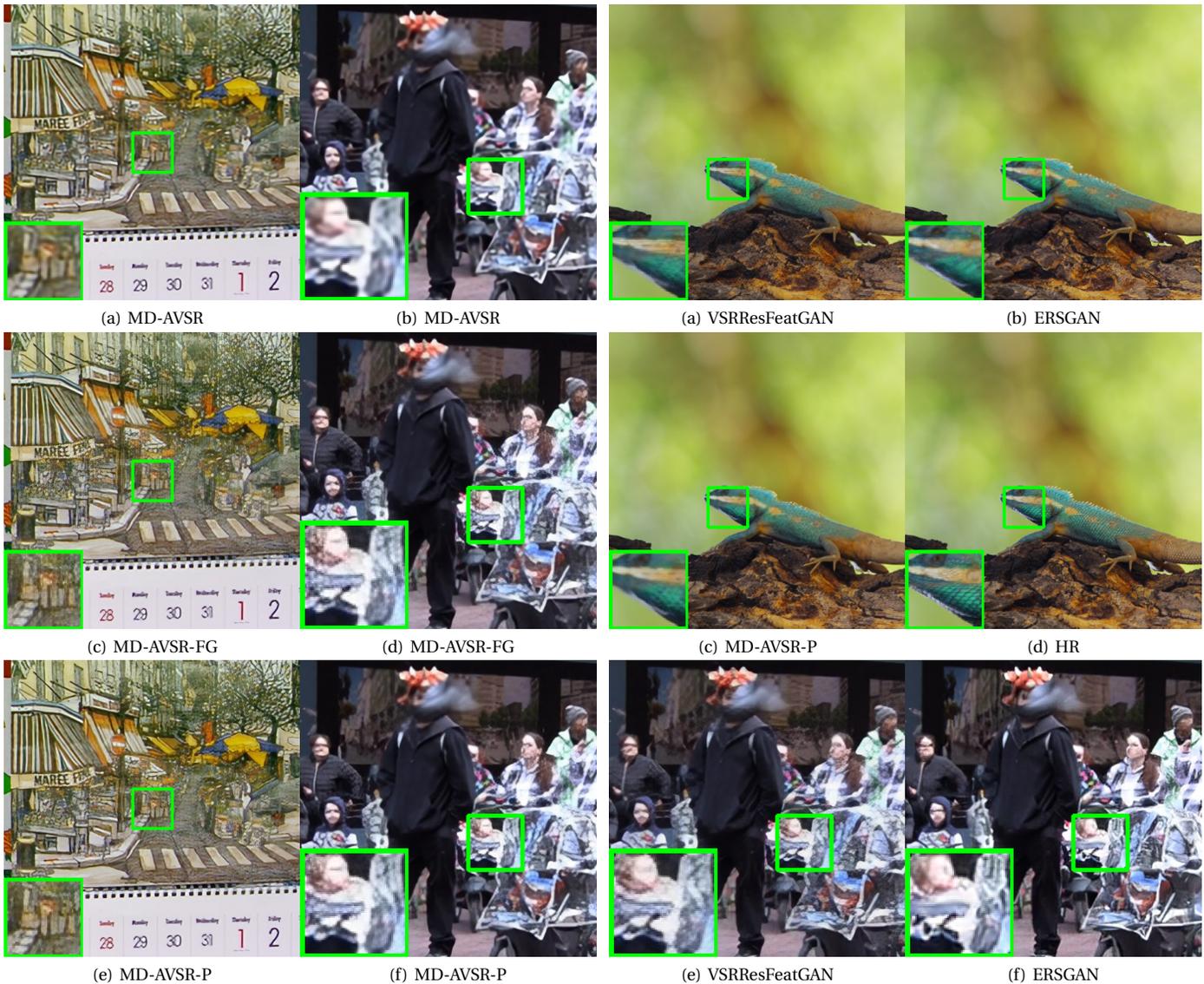

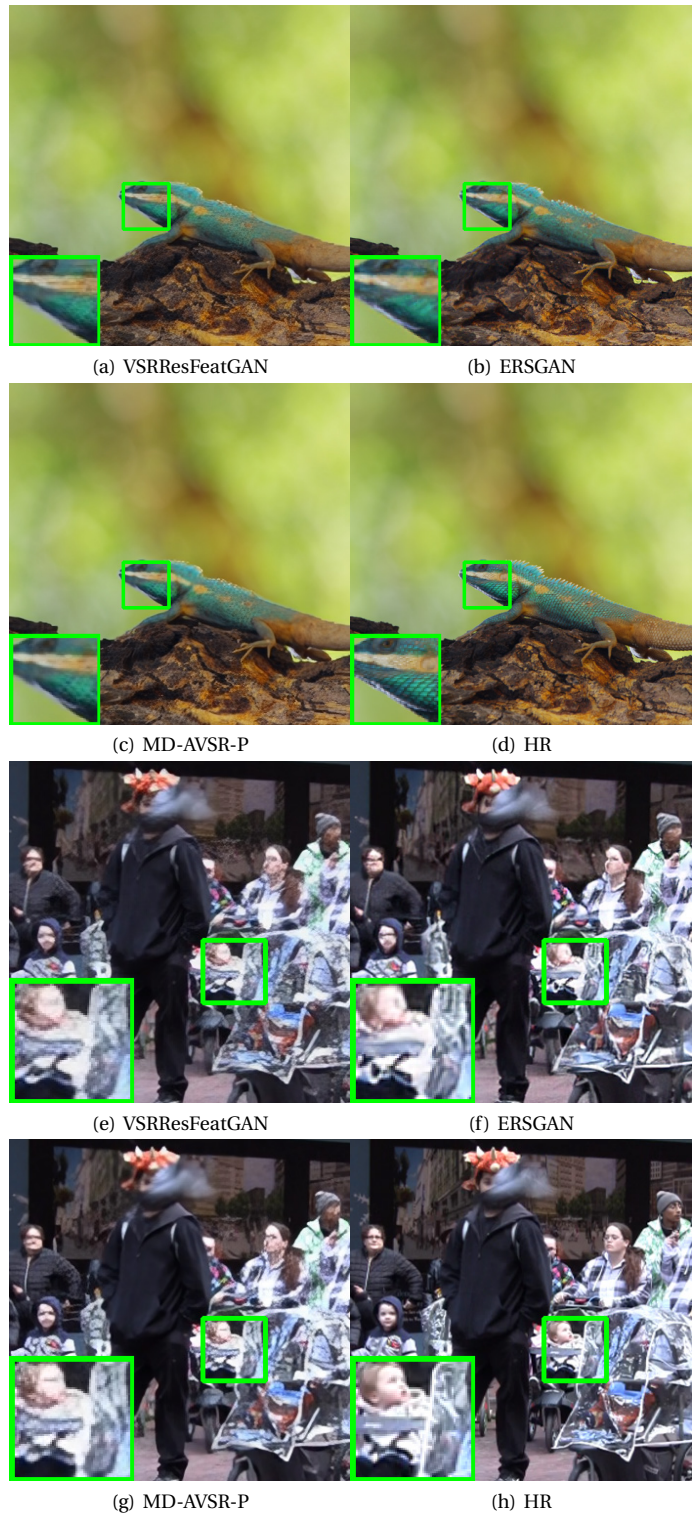

Fig. 7: Qualitative comparison between MD-AVSR, MD-AVSR-FG and MD-AVSR-P for factor 4 with bicubic downsampling. We can see that MD-AVSR-P does not produce artifacts like MD-AVSR-FG and produces frames that look much sharper than MD-AVSR.

Table II also shows that MD-AVSR-PY8 outperforms MD-AVSR-P in terms of SSIM and Perceptual Distance significantly, although it suffers from a minor decrease in PSNR. This increase shows the importance of using a very diverse dataset during training for GAN models in contrast to experiments carried out with the non GAN model MD-AVSR, where the training with this new dataset did not produce results different enough to be significative for any factor. These results show that GAN-based VSR models require more data that other CNN. Fig 9 shows a qualitative comparison between MD-AVSR-P and MD-AVSR-PY8. It can be seen that MD-AVSR-PY8 is able to produce more detailed and realistic looking images.

Fig. 8: Comparison between VSRResFeatGAN, ERSGAN and MD-AVSR-P for factor 4 with bicubic downsampling.

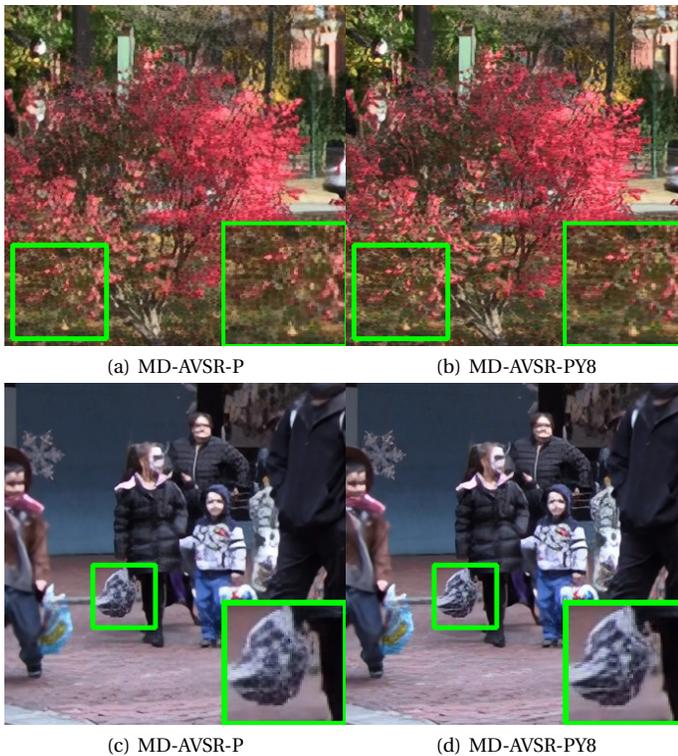

(a) MD-AVSR-P        (b) MD-AVSR-PY8

(c) MD-AVSR-P        (d) MD-AVSR-PY8

Fig. 9: Comparison between MD-AVSR-P and MD-AVSR-PY8 for factor 4 with bicubic downsampling.

## V. Conclusions

In this work we have first introduced a multiple degradation Video Super-Resolution approach that explicitly utilizes the LR image formation model as an input to the network. The model, named MD-AVSR, has been trained with MSE only. The experiments show that MD-AVSR outperforms current state-of-the-art methods in terms of PSNR and SSIM for both multiple degradation and bicubic degradation only settings. We have then proposed a GAN-based approach that uses a new perceptual loss combining an adversarial loss, a feature loss, and a space smoothness constraint, the model has been named MD-AVSR-P. This method improves the quality of the super resolved frames without the introduction of noticeable high frequency artifacts. The results show that it outperforms current state-of-the-art methods in terms of perceptual quality and in all metrics GAN methods trained without using the proposed perceptual loss. Finally, we use a much more diverse dataset created from a subset of the YouTube-8M dataset to train MD-AVSR-P and show that the new MD-AVSR-PY8 obtains significantly better results in terms of perceptual quality.